\documentclass{article}
\usepackage{ijcai26}

\usepackage{times}
\usepackage{soul}
\usepackage{url}
\usepackage[hidelinks]{hyperref}
\usepackage[utf8]{inputenc}
\usepackage[small]{caption}
\usepackage{graphicx}
\usepackage{amsmath}
\usepackage{amsthm}
\usepackage{booktabs}
\usepackage{algorithm}
\usepackage{algorithmic}
\usepackage[switch]{lineno}

\usepackage{natbib}
\usepackage{wrapfig}
\usepackage{tabularx}
\usepackage{multirow}
\usepackage{subcaption}
\usepackage{xcolor}
\usepackage{makecell}

\newtheorem{lemma}{Lemma}  
\newtheorem{proposition}{Proposition}  

\title{CAPER: Constrained and Procedural Reasoning for Robotic \\Scientific Experiments}

\author{
Jinghan Yang$^1$
\and
Jingyi Hou$^{1*}$\and
Xinbo Yu$^1$\And
Wei He$^{1,2}$
\And
Yifan Wu$^1$\\
\affiliations
$^1$University of Science and Technology Beijing\\
$^2$Beijing Information Science and Technology University\\
$^*$houjingyi@ustb.edu.cn\\
}

\begin{document}

\maketitle

\begin{abstract}
Robotic assistance in scientific laboratories requires procedurally correct long-horizon manipulation, reliable execution under limited supervision, and robustness in low-demonstration regimes. Such conditions greatly challenge end-to-end vision-language-action (VLA) models, whose assumptions of recoverable errors and data-driven policy learning often break down in protocol-sensitive experiments. We propose CAPER, a framework for Constrained And ProcEdural Reasoning for robotic scientific experiments, which explicitly restricts where learning and reasoning occur in the planning and control pipeline. 
Rather than strengthening end-to-end policies, CAPER enforces a responsibility-separated structure: task-level reasoning generates procedurally valid action sequences under explicit constraints, mid-level multimodal grounding realizes subtasks without delegating spatial decision-making to large language models, and low-level control adapts to physical uncertainty via reinforcement learning with minimal demonstrations. 
By encoding procedural commitments through interpretable intermediate representations, CAPER prevents execution-time violations of experimental logic, improving controllability, robustness, and data efficiency. Experiments on a scientific workflow benchmark and a public long-horizon manipulation dataset demonstrate consistent improvements in success rate and procedural correctness, particularly in low-data and long-horizon settings.
\end{abstract}

\section{Introduction}


Robotic assistance in scientific laboratories aims to support researchers during the research-and-development (R\&D) stage by offloading repetitive and time-consuming experimental procedures, such as material synthesis, chemical preparation, and standardized testing.
While recent systems have demonstrated reliable automation in highly structured and fully specified laboratory pipelines \citep{pyzer2022accelerating,wang2024enhancing,ramos2025review}, real-world R\&D experiments impose requirements that fundamentally differ from those of routine or closed-form automation.
Scientists often specify only high-level experimental goals, leaving critical procedural details (e.g., step ordering, prerequisite conditions, and irreversible operations) underspecified.
As a result, success in scientific experimentation is determined not merely by task completion, but by strict adherence to experimental protocols.
Moreover, experimental workflows in R\&D settings are often executed under limited supervision, with sparse, weak, or biased demonstrations that reflect procedural conventions rather than action trajectories.
These characteristics make long-horizon robotic manipulation in scientific laboratories uniquely challenging, demanding procedural correctness, robustness under weak supervision, and reliability over extended action sequences.

Recent advances in large language models (LLMs) have demonstrated strong capabilities in task decomposition and logical reasoning \citep{ha2023scaling,dalal2024plan,zhou2024isr,guo2024castl}, while vision-language-action (VLA) models \citep{DBLP:journals/corr/abs-2405-14093,intelligence2025pi_,zhao2025cot} attempt to unify perception and reasoning in an end-to-end manner. 
Despite their success on household and industrial benchmarks, these methods might face limitations in aligning with R\&D-stage scientific experimentation.
End-to-end VLAs implicitly rely on the recoverability of execution errors, dense trajectory supervision, and the ability to absorb task logic into latent action representations.
In scientific experiments, these assumptions break down, as errors are often irreversible and procedural validity cannot be learned reliably from sparse demonstrations.
As task horizons grow and experimental procedures become compositional, the space of valid action sequences expands combinatorially, making purely data-driven generalization expensive.
This limitation echoes broader findings on systematic generalization \citep{DBLP:conf/icml/LakeB18,DBLP:conf/iclr/BahdanauMNNVC19}, and is further exacerbated by recent evidence that task-specific fine-tuning of VLA models can bias policies toward superficial goal completion while neglecting underlying state and protocol correctness \citep{DBLP:conf/nips/LiCC24}. Together, these observations suggest that the challenges faced by end-to-end VLA models in scientific domains are not incidental, but structural.
Such evidence reveals the inherent shortcomings of purely end-to-end learning in robotic scientific experiments.

\begin{figure*}[t]
\centerline{\includegraphics[width=0.8\linewidth]{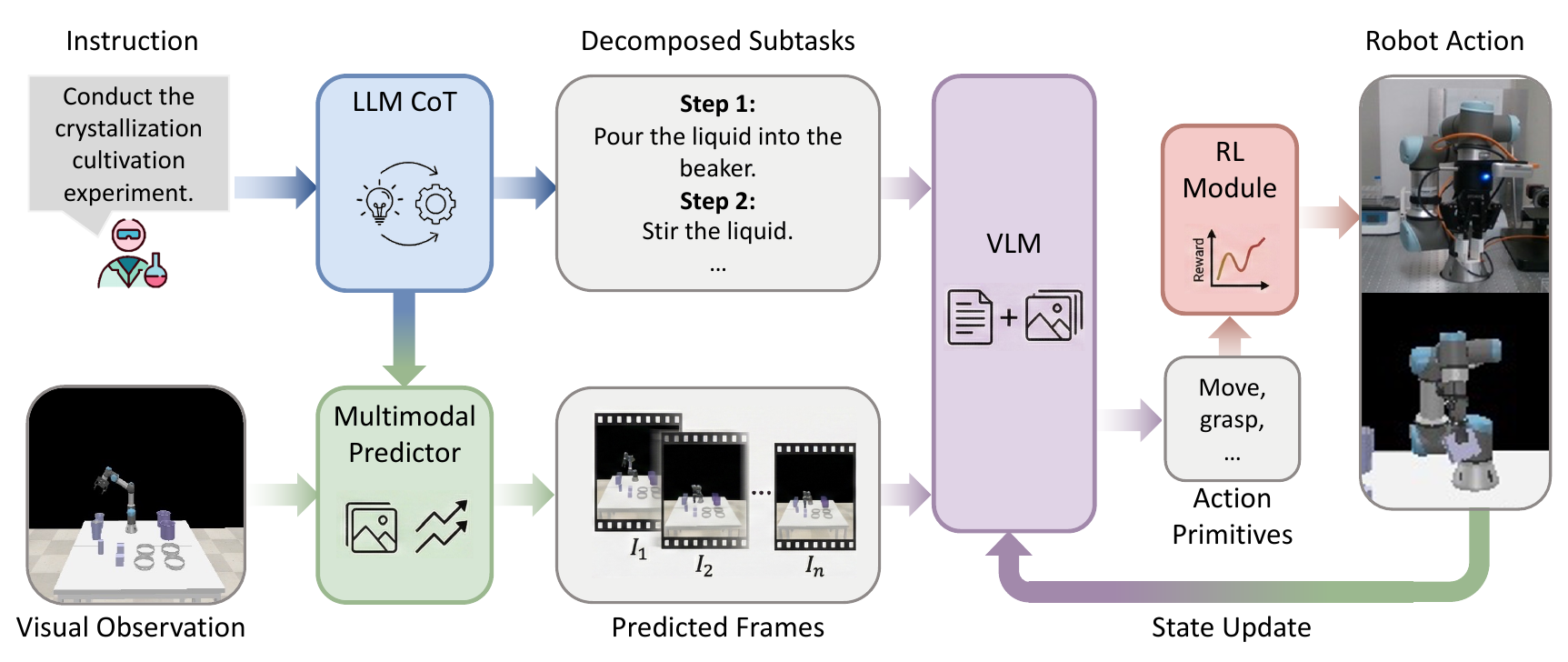}}
\caption{ 
Overview of the CAPER framework.
CAPER decomposes long-horizon tasks into modular stages. 
First, an LLM generates procedurally valid sequence of high-level subtasks in symbolic space, without access to visual or control signals.
A multimodal planner, composed of a multimodal predictor and a VLM, then grounds each subtask by conditioning on the current and predicted goal images, producing sequences of action primitives. Finally, a low-level controller executes these primitives in the environment, enabling safe and consistent task completion.
}
\label{fig:overview}
\end{figure*}


These limitations suggest rethinking how learning and reasoning are structured in long-horizon robotic manipulation for scientific R\&D experiments. 
Rather than strengthening end-to-end policies, learning and reasoning should be explicitly constrained.
In scientific workflows, procedural logic and perceptual uncertainty often arise at different levels. 
Deferring procedural reasoning to latent action representations couples these factors, complicating error analysis and corrective intervention.
Instead, procedural reasoning should be treated as an explicit commitment made prior to execution, narrowing the space of admissible actions and shielding downstream modules from semantic ambiguity. 
Instead, making procedural reasoning explicit prior to execution allows the action space to be constrained in advance, reducing ambiguity during downstream execution.
Motivated by this observation, we propose CAPER (Constrained And ProcEdural Reasoning) for robotic Scientific Experiments, a framework that enforces a responsibility-separated structure across task-level reasoning, mid-level grounding, and low-level control. 
By explicitly encoding procedural commitments through interpretable intermediate representations, CAPER prevents execution-time violations of experimental logic and enables each component to focus on the uncertainties it is best suited to handle.

Specifically, CAPER organizes long-horizon robotic manipulation into three decoupled stages that reflect the structure of scientific workflows. 
Task-level reasoning procedurally valid action sequences under explicit constraints, resolving ordering and prerequisite relations before conduct. 
Mid-level grounding realizes each subtask through multimodal perception without delegating spatial decision-making to LLMs.
Low-level control then adapts execution to physical uncertainty through reinforcement learning, without relying on explicit trajectory supervision.
This responsibility-separated design allows procedural logic, perceptual grounding, and physical control to be handled at appropriate levels of abstraction. 
We evaluate CAPER on a scientific workflow benchmark designed to emphasize procedural validity, as well as on a public long-horizon manipulation dataset. 
Across both settings, CAPER achieves consistent improvements in task success and procedural correctness, with particularly strong gains in low-data and long-horizon regimes.

\section{Related Work}

\paragraph{Robot manipulation learning}
Robotic manipulation has traditionally relied on symbolic planning and control methods such as trajectory optimization \citep{parr1997reinforcement,kaelbling2011hierarchical}, which work well in structured settings but lack flexibility and generalization. Recent advances in LLMs \citep{radford2018improving,touvron2023llama} have inspired their use in robotic skill learning \citep{ahn2022can}, programmatic task decomposition \citep{singh2023progprompt}, and in combination with RL for low-level control \citep{dalal2024plan}, though these approaches are often limited by unstable execution and poor systematic generalization. VLA models aim to unify perception, reasoning, and control \citep{DBLP:journals/corr/abs-2505-22159,DBLP:conf/hri/SautenkovYLMTAC25,DBLP:journals/corr/abs-2503-20384}, 
but purely end-to-end training struggles to scale as task compositions grow combinatorially, since task-level constraints and procedural dependencies must be implicitly absorbed into action distributions.
CoT-VLA \citep{zhao2025cot} and HAMSTER \citep{DBLP:conf/iclr/0038DZJMG0F00025} move toward hierarchical structures by incorporating CoT reasoning or coarse-to-fine path planning, yet they remain tightly coupled through trajectory-level supervision, where intermediate reasoning and execution are co-adapted rather than independently optimized.
In contrast, our proposed CAPER explicitly disentangles task decomposition, multimodal prediction, and low-level control into heterogeneous modules. 
This modularity allows independent optimization without trajectory-level supervision, enabling controllable procedural reasoning and robust compositional recombination under domain shifts.

\paragraph{Autonomous experimental systems for science}
Prior works have focused on autonomous scientific discovery by combining high-throughput experimentation and AI-driven planning. 
For example, \citep{zhao2021discovery} leverage automated platforms to explore the stability of perovskites under varying conditions. 
\citep{coley2019robotic} develop an organic synthesis platform using robotic arms and modular components to automate complex chemical processes. 
\citep{burger2020mobile} propose a mobile chemical robot that performs hundreds of experiments autonomously to discover a novel hydrogen-production catalyst.
These systems typically assume tightly integrated, domain-specific platforms and predefined experimental pipelines, limiting their flexibility for open-ended laboratory workflows.
Our work focuses instead on experimental assistance during the R\&D process to help researchers 
perform tedious lab tasks more efficiently. 

\begin{figure} 
    \centering
    \includegraphics[width=\linewidth]{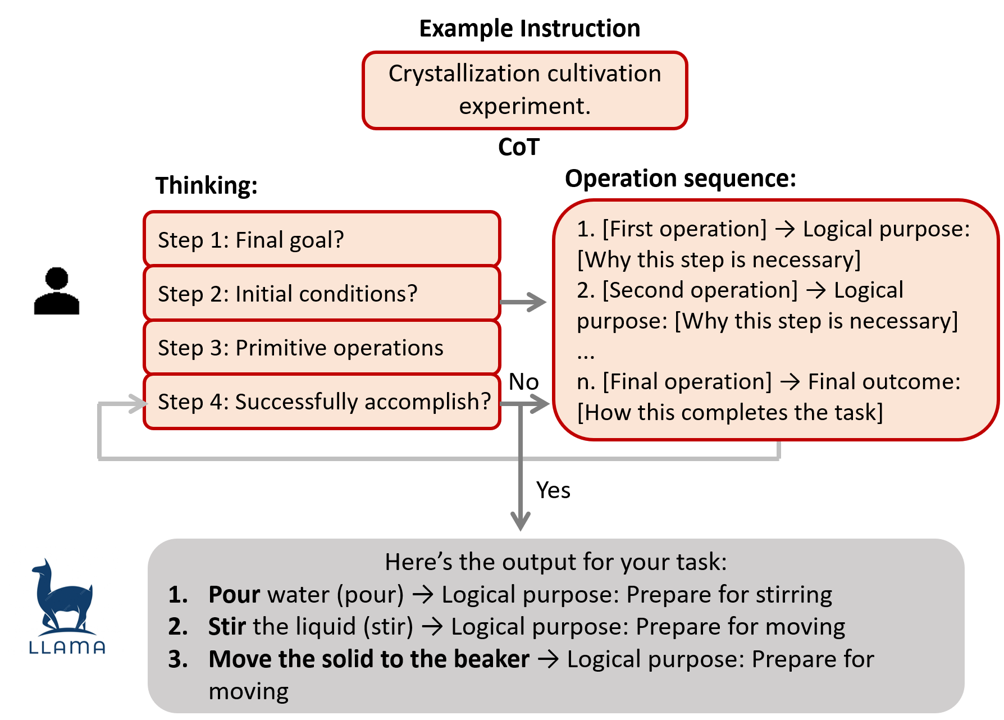}
    \caption{An example of the task-level planner using CoT reasoning. 
The LLM is guided through a 4-step process: 
(1) clarify the goal, 
(2) identify required objects and conditions, 
(3) decompose into basic operations, and 
(4) validate task completion.
}
    \label{fig:cot}
\end{figure}

\section{Method}


\subsection{Pipeline overview}\label{AA}



Long-horizon robotic planning requires reasoning, visual grounding, and reliable skill execution, which are challenging to achieve within a single end-to-end model. 
To address this, we introduce a constrained and procedural reasoning framework (CAPER) that disentangles these stages (Fig.~\ref{fig:overview}).

\subsection{Task-level planner}

We model the task-level planner as a symbolic decomposition process, operating entirely in the language space and decoupled from perception or control signals. 
As shown in Fig. \ref{fig:cot}, given a high-level goal $G$ (e.g., ``crystallization cultivation experiment.
''), we use CoT reasoning with the Meta-Llama-3.1-8B-Instruct model to generate a sequence of semantically coherent subtasks 
\begin{equation}
\mathcal{S} = \{s_1, s_2, \ldots, s_T\}, \quad s_t \in \mathcal{L},
\end{equation}
where $\mathcal{L}$ denotes the symbolic instruction space (natural-language actions such as ``add reagent A'').  

To construct $\mathcal{S}$, we employ CoT prompting, which decomposes reasoning into four stages:
\[
G \xrightarrow{\text{(1) interpret}} g 
\xrightarrow{\text{(2) prerequisites}} \mathcal{P} 
\xrightarrow{\text{(3) decompose}} \mathcal{S}
\xrightarrow{\text{(4) validate}} \mathcal{S}^*,
\]
where $g$ denotes the interpreted core objective, $\mathcal{P}$ is the set of prerequisite conditions, and $\mathcal{S}^*$ is the optimized subtask sequence after logical validation.

To improve reliability, we introduce a verification–correction loop applied iteratively:
\begin{equation}
\mathcal{S}^{(k+1)} = \Phi\big(\mathcal{S}^{(k)}\big), \quad 
\Phi = f_{\text{val}} \circ f_{\text{cons}} \circ f_{\text{corr}},
\end{equation}
where $f_{\text{val}}$ performs subgoal validation, $f_{\text{cons}}$ checks temporal/causal/physical constraints, and $f_{\text{corr}}$ applies corrections (see detailed description in Appendix~\ref{app:self-ver}). 
This process continues until $\mathcal{S}^{(k)}$ converges to a logically consistent plan $\mathcal{S}^\ast$.
This mechanism operates entirely within the LLM without external labels, leveraging its inherent reasoning abilities to ensure semantic correctness. 
As the task-level planner is performed in the symbolic space of language, it is decoupled from direct visual inputs. 

The output $\mathcal{S}^\ast$ resides purely in the symbolic language space and is therefore independent of low-level visual observations. Formally, we define a two-stage factorization of the overall policy:
\begin{equation}
\pi(a_t \mid o_t, G) 
= \pi_{\text{exec}}(a_t \mid o_t, \mathcal{S}^\ast), \quad 
\mathcal{S}^\ast = \pi_{\text{plan}}(G),
\end{equation}
where $\pi_{\text{plan}}$ maps high-level goals $G$ into symbolic plans, and $\pi_{\text{exec}}$ grounds these symbolic subtasks into action primitives $a_t$ conditioned on visual observations $o_t$, which will be described in detail in the following section w.r.t the multimodal planner.  

This factorization highlights the modularity of our framework: the task-level planner is responsible for semantic reasoning in symbolic space, while mid-level multimodal planner and low-level controller modules handle grounding and execution. 
Such a separation improves robustness and reusability, especially in complex, sensor-rich scientific environments.
We provide a detailed discussion on the feasibility of the decoupling manner in Appendix~\ref{app:decouple}.

\subsection{Mid-level multimodal planner}


The mid-level multimodal planner bridges high-level task decomposition and low-level control by grounding each subtask in perceptual observations. 
It consists of two components: (1) a multimodal predictor that provides execution-relevant visual context by predicting future scene states, and (2) a VLM that maps grounded subtasks to structured action primitives that can be executed by the controller.



\subsubsection{Visual prediction}

The multimodal predictor is based on a conditional diffusion model~\citep{du2024learning} and is used to estimate short-horizon future visual states during subtask execution.
Given the current observation and a subtask description, the model predicts a sequence of intermediate frames that reflect how the scene is likely to evolve if the subtask is carried out.
These predicted frames are not used to choose actions, but to provide visual context that supports grounding and execution.

Task instructions are encoded by a frozen CLIP encoder, while a UNet backbone extracts visual features from the current frame.
We fuse the two streams via cross-attention, so the predictor conditions on both the scene and the subtask semantics.

We use a conditional diffusion formulation.
Let $x_0$ denote the clean future frame (or a future-frame sequence element) and $x_t$ the noisy sample at diffusion step $t$.
The forward noising process is
\begin{equation}
q(x_t \mid x_{t-1})=\mathcal{N}\!\left(x_t;\sqrt{1-\beta_t}\,x_{t-1},\,\beta_t \mathbf{I}\right),
\end{equation}
with a predefined noise schedule $\{\beta_t\}_{t=1}^T$.
In the reverse process, the model predicts the noise and denoises step-by-step:
\begin{equation}
x_{t-1}=\frac{1}{\sqrt{1-\beta_t}}
\left(x_t-\frac{\beta_t}{\sqrt{1-\bar{\alpha}_t}}\,
\epsilon_\theta(x_t,t)\right)+\sigma_t z,
\end{equation}
where $z\sim\mathcal{N}(0,\mathbf{I})$ and $\epsilon_\theta$ is parameterized by a UNet conditioned on the current observation and the CLIP embedding.

Given the current frame $i_0$ and subtask text $l$, the predictor outputs a short sequence of future frames
\begin{equation}
P_\theta(i_0,l)\rightarrow \{ \hat{i}_1,\hat{i}_2,\ldots,\hat{i}_K \},
\label{eq:predictor}
\end{equation}
which provides visual context for grounding.
We train the predictor with the standard noise-prediction objective and apply SNR-based reweighting~\citep{ho2020denoising}
and use gradient clipping for stability.

For scientific experiment scenes, we train on trajectories collected in a simulated laboratory environment.
We crop low-information regions, use $256\times256$ inputs, and sample $K{=}8$ frames per sequence.
By predicting likely future configurations, the model helps expose spatial conflicts (e.g., potential collisions) and improves the robustness of downstream grounding and execution.

\begin{figure} 
    \centering
    \includegraphics[width=\linewidth]{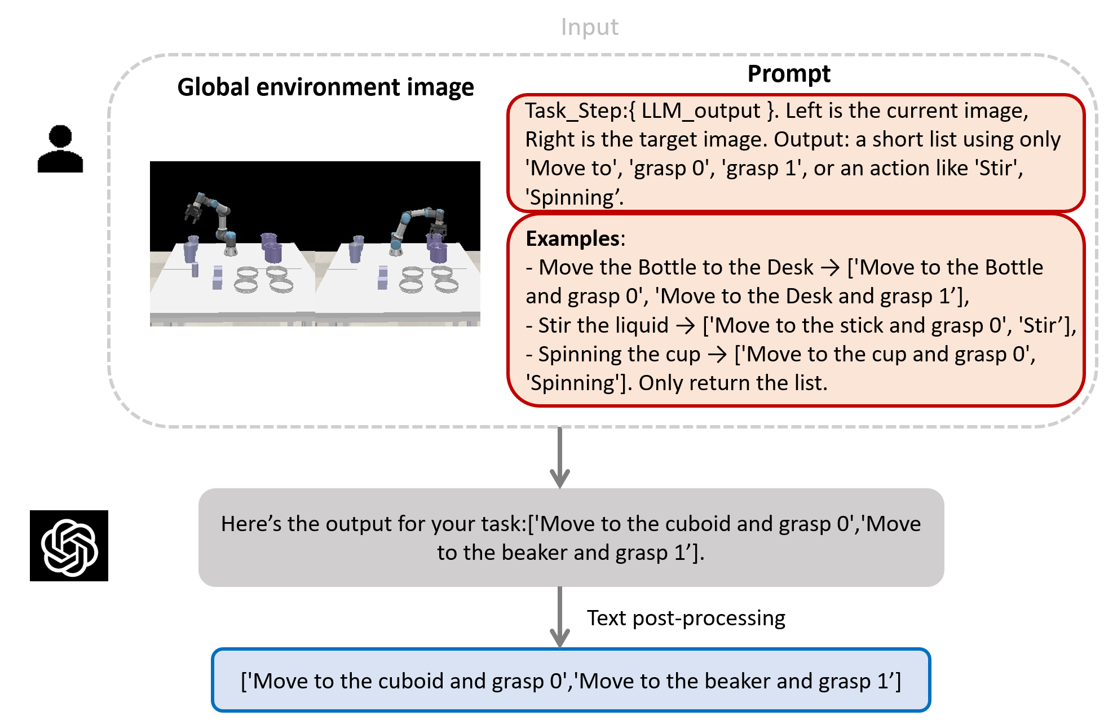}
    \caption{An example prompt used for the VLM.}
    \label{fig:prompt}
\end{figure}

\subsubsection{Action primitive generation}

Given a symbolic subtask and visual observations, we convert the subtask into a sequence of executable action primitives.
We use a vision-language model (GPT-4o) to map each subtask to a small set of predefined action primitives commonly used in laboratory settings: \texttt{move}, \texttt{grasp}, \texttt{pour}, and \texttt{stir}.
To reduce ambiguity, we adopt a structured prompt format (Fig.~\ref{fig:prompt}) that conditions the VLM on paired visual inputs (the current observation and a predicted future frame) together with the subtask description.

Most subtasks are realized through combinations of \texttt{move} and \texttt{grasp}, with sequential execution represented by simple concatenation (e.g., ``\texttt{move} and \texttt{grasp}'').
Operations such as \texttt{pour} and \texttt{stir} are treated as atomic actions and are invoked directly when required.
Each \texttt{move} primitive is expressed as \texttt{move to <target>}, and each \texttt{grasp} as \texttt{grasp 1/0}, indicating grasp or release.

The VLM outputs a structured sequence of action primitives that serves as an interpretable and robot-compatible intermediate representation.
This representation provides a simple interface to the low-level controller and supports downstream reinforcement learning, where execution policies are refined through interaction with the environment.




\subsection{Low-level controller}

Recent diffusion-based and ACT-based VLAs show strong trajectory generation ability, but often suffer from fragile visual grounding when perception is implicitly coupled with action generation.
In end-to-end formulations, grounding errors directly translate into execution failures and repeated attempts, without an explicit interface for inspection or correction.
CAPER avoids this issue by resolving task reasoning and visual grounding upstream, and restricting the low-level controller to executing fixed action primitives.

To execute the action primitives, we use a continuous-control policy that maps observations to joint-level commands.
While DDPG~\citep{DBLP:journals/corr/LillicrapHPHETS15} is one possible choice, other continuous controllers can be used.
Reinforcement learning is applied only at this level to improve primitive execution under local feedback, without reintroducing perceptual or task-level uncertainty.

The reward function balances task progress, success, and safety:
\begin{equation}
r = r_\text{move} + r_\text{grasp} - r_\text{collision},
\end{equation}
where $r_\text{move}$ encourages approaching the target, $r_\text{grasp}$ rewards successful grasps, and $r_\text{collision}$ penalizes collisions.

To support long-horizon tasks, the environment is reset only at task initialization or completion, maintaining continuity across subtasks.

\subsection{Simulation setup}

\begin{table*}[tbp]
\small 
\caption{Training tasks and prompt templates of the scientific experiment dataset.}
\centering
\setlength{\tabcolsep}{6pt}
\begin{tabularx}{\linewidth}{l X X c c}
\toprule
Task & Description & Prompt template & Environment & \#Training data \\ 
\midrule

Pick\_place &
Basic object relocation. &
\makecell[l]{
1. Move $<$object$>$ to $<$object$>$. \\
2. Put $<$object$>$ on $<$object$>$. \\
3. Place $<$object$>$ next to $<$object$>$. \\
$\cdots$
} &
Simulation &
88 \\

Pour &
Single-step pouring of liquid. &
\makecell[l]{
1. Pour $<$liquid$>$. \\
2. Pour $<$liquid$>$ into $<$container$>$. \\
3. Pour $<$liquid$>$ from $<$container$>$. \\
$\cdots$
} &
Simulation &
25 \\

Stir &
Stir the liquid in a container. &
\makecell[l]{
1. Stir $<$liquid$>$. \\
2. Stir $<$liquid$>$ in $<$container$>$. \\
3. Perform stirring. \\
$\cdots$
} &
Simulation &
27 \\

Mix &
Multi-step reactant mixing experiment. &
\makecell[l]{
1. Conduct reactant mixing experiment. \\
2. Perform multi-step mixing. \\
3. Combine $<$liquid$>$ and $<$liquid$>$. \\
$\cdots$
} &
Simulation &
50 \\

Crystallize &
Crystallization cultivation experiment. &
\makecell[l]{
1. Conduct crystallization. \\
2. Grow $<$object$>$ from $<$solution$>$. \\
3. Perform crystallization cultivation. \\
$\cdots$
} &
Simulation &
73 \\

Weigh &
Weigh the liquid content within the container. &
\makecell[l]{
1. Weigh $<$object$>$. \\
2. Weigh $<$object$>$ with the balance. \\
$\cdots$
} &
Real-world &
27 \\

Shake &
Load the container onto the shaker. &
\makecell[l]{
1. Load $<$container$>$ onto the shaker. \\
2. Shake $<$container$>$. \\
$\cdots$
} &
Real-world &
25 \\

\bottomrule
\end{tabularx}
\label{tab:dataset}
\end{table*}

We build the simulation environment using CoppeliaSim (V-REP), designed for scientific experimental scenarios.
A UR3 robotic arm is used as the execution platform.
The environment includes common laboratory apparatus such as beakers and Petri dishes, as well as graspable objects including cylinders, cubes, cups, and sticks, which abstract different categories of experimental items.

The simulator provides collision detection and distance sensing. Low-level controllers can access object states through the environment interface to compute rewards during training.
Based on this environment, we collect a dataset for scientific manipulation tasks, covering both long-horizon and extra-long-horizon experimental procedures.
Instead of assigning a single fixed instruction to each task, we construct a pool of prompt templates for every task category.
Each pool contains semantically equivalent but lexically diverse instructions generated by an LLM, increasing linguistic variability while preserving task intent.

The dataset includes tasks such as object relocation, pouring, stirring, multi-step reactant mixing, and crystallization cultivation.
For each task instance, the dataset records the execution trajectories required to complete the target procedure.
The number of training samples for each task is summarized in Table~\ref{tab:dataset}.
This dataset is sufficient for training the multimodal predictor to produce stable and accurate visual predictions.

For the evaluation of each task, multiple instructions are sampled from the corresponding prompt-template pool and paired with different object combinations to form 20 distinct input prompts.
For each prompt, we further sample 50 randomized initial scenes with varying object poses and spatial layouts, resulting in 1,000 evaluation trials per task.

\section{Experiments} 
\subsection{Experimental setup}

All experiments are conducted on a single NVIDIA GeForce RTX 4090 GPU for both training and inference.
Due to GPU memory constraints, we use a batch size of 1, where each batch contains 8 consecutive robot motion frames from a single task instance.
All input images are resized to a resolution of $256 \times 256$.
For low-level control, we employ DDPG, with a replay buffer size of 4.


\subsection{Visualization}

\begin{figure}[tbp]
\centering
\includegraphics[width=0.8\linewidth]{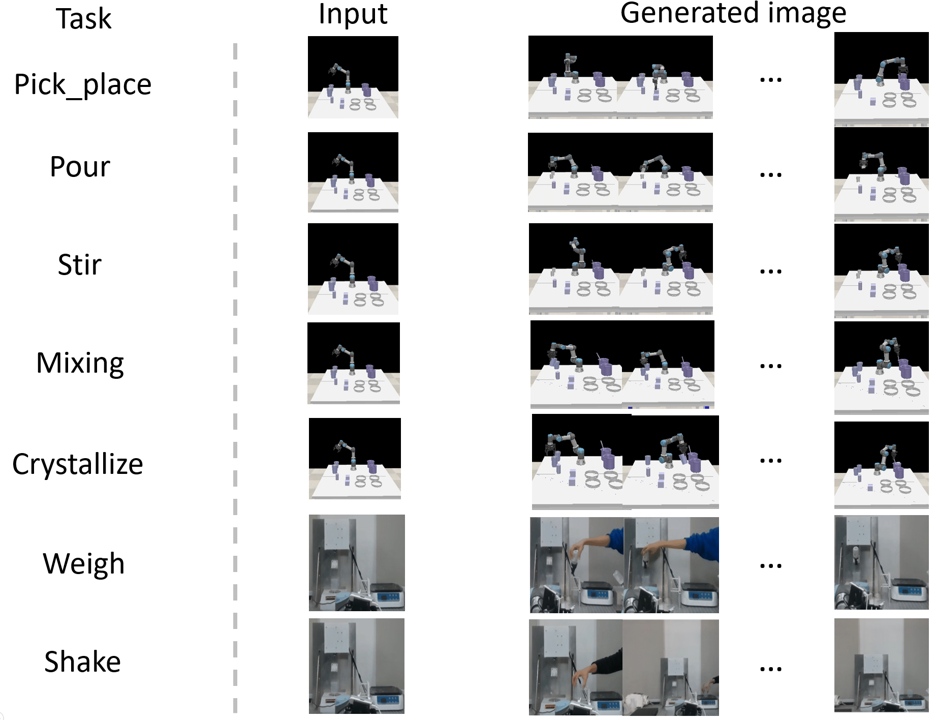}
\caption{Examples of predicted future visual frames generated by the multimodal predictor under different task conditions.}
\label{fig:example}
\end{figure}

\begin{figure}[tbp]
\centering
\includegraphics[width=\linewidth]{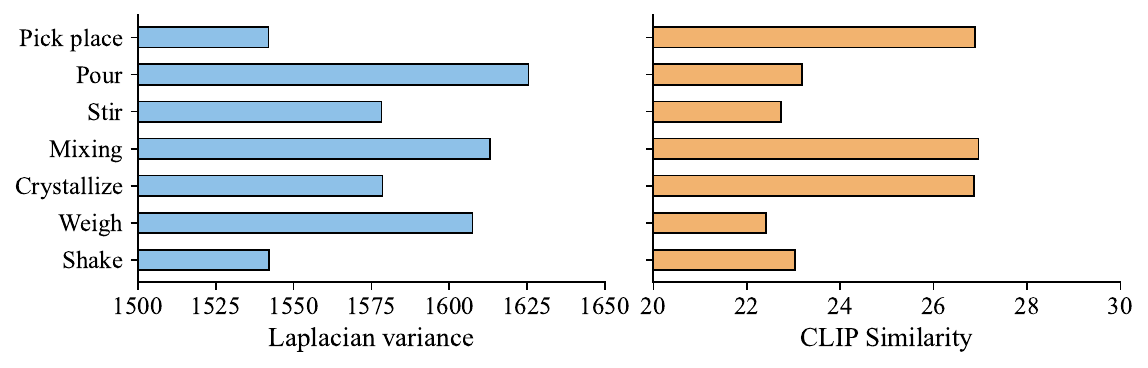}
\caption{Visual clarity and semantic relevance of predicted images.}
\label{fig: clarity}
\end{figure}


To evaluate the quality of future frames generated by the multimodal predictor, we measure bothvisual clarity and semantic alignment using standard quantitative metrics:

\noindent\textbf{Laplacian Variance} is used to quantify image sharpness by computing the variance of the Laplacian response for each predicted frame.

\noindent\textbf{CLIP Similarity} quantifies the semantic alignment between task instructions and predicted images.
We extract embeddings using a pretrained CLIP model and compute cosine similarity between the instruction text and the corresponding predicted frames.

Results are shown in Fig. \ref{fig: clarity}.
The predicted frames consistently achieve Laplacian variance values above 1,500, indicating sufficient visual sharpness to preserve motion boundaries and object contours.
In addition, CLIP similarity scores remain above 21, suggesting strong semantic alignment between task descriptions and predicted visual content.
Such alignment supports stable subtask grounding by providing task-consistent visual context.

Examples of predicted future frames for different tasks are shown in Fig.~\ref{fig:example}.

\subsection{Experimental results on the scientific experiment dataset (simulation)}

We test the success rates of CAPER and other methods using the benchmark we designed.
In approaches that only rely on RL model, the DDPG model exhibits near-zero success rates on complex, long-horizon tasks due to the absence of task decomposition and planning mechanisms. 
Its performance is nearly indistinguishable from random trial-and-error, making meaningful comparison infeasible.
The success rates of different methods are shown in Table \ref{tab:success rate of different methods}.

\noindent\textbf{LLM+RL (LLaMA-3.1-8B, w/o CoT).} 
We remove the multimodal predictor (MP) and replace the mid-level VLM with LLaMA-3.1-8B, without CoT prompting.
Subtasks are generated directly from textual task and environment descriptions and executed by the RL controller.
This baseline evaluates the limitation of language-only task decomposition without visual grounding or structured reasoning.

\noindent\textbf{VLM+RL (LLaMA3.2, w/o CoT).}  
We introduce visual input by replacing the LLM with a VLM.
This setting evaluates the contribution of visual information to task decomposition in the absence of CoT reasoning.

\noindent\textbf{LLM+RL (GPT-4o).}
We use GPT-4o with CoT prompting for task decomposition, but without visual input.
This baseline isolates the effect of strong language reasoning without visual grounding.

\noindent\textbf{VLM+RL (GPT-4o).} 
We remove both the multimodal predictor and the task-level planner from CAPER, relying on a VLM to generate subtasks directly from visual observations.
Compared with the LLaMA3.2-based VLM+RL baseline, this setting highlights the effectiveness of our structured prompting, while the performance drop indicates the importance of explicit grounding and planning.

\noindent\textbf{MP+VLM+RL (GPT-4o).} 
We remove only the task-level planner from CAPER to assess its role.
Performance differences are minor for short-horizon tasks, but become pronounced for long-horizon tasks such as {mixing} and {crystallize}, demonstrating the necessity of explicit task-level planning.


\noindent\textbf{CAPER.}
CAPER achieves the best overall performance on the scientific experiment benchmark.
It performs strongly on relatively short-horizon tasks such as {pick\_place}, while showing a moderate performance drop on {pour} and {stir}, primarily due to grasp instability caused by thin tools and challenging end-effector poses, which increase the likelihood of slippage during execution.
For long-horizon tasks including {mixing} and {crystallize}, success rates decrease further as visual ambiguity and procedural complexity accumulate over time.
Nevertheless, CAPER consistently outperforms all baselines, demonstrating the necessity and effectiveness of integrating task-level planning, explicit grounding, and learned execution.



\begin{table*}[tbp]
\centering
\setlength{\tabcolsep}{10pt} 
\begin{tabularx}{\linewidth}{lcccccc}
\toprule
Task &
\makecell[c]{LLM+RL \\ \scriptsize{(Llama-3.1-8B w/o CoT)}} &
\makecell[c]{MLM+RL \\ \scriptsize{(LLaMA3.2 w/o CoT)}} &
\makecell[c]{LLM+RL \\ \scriptsize{(GPT-4o)}} &
\makecell[c]{MLM+RL \\ \scriptsize{(GPT-4o)}} &
\makecell[c]{MP+MLM+RL \\ \scriptsize{(GPT-4o)}} &
\makecell[c]{CAPER } \\
\midrule
Pick\_place   & 39.6 & 45.8 & 70.4 & 78.9 & 79.3 & 80.3 \\
Pour          & 34.8 & 44.7 & 54.6 & 62.6 & 62.4 & 64.5 \\
Stir          & 30.2 & 39.6 & 34.8 & 70.5 & 70.8 & 74.4 \\
Mix           & 23.3 & 35.5 & 23.8 & 43.2 & 43.6 & 52.3 \\
Crystallize   & 11.6 & 12.9 & 12.4 & 21.0 & 21.5 & 30.1 \\
\bottomrule
\end{tabularx}
\caption{Success rate (\%) of different methods on the scientific experiment dataset.}
\label{tab:success rate of different methods}
\end{table*}

\begin{table*}[tbp]
\centering
\setlength{\tabcolsep}{18pt} 
\begin{tabularx}{\linewidth}{lccccc}
\toprule
Task & PerAct & OpenVLA & CoT-VLA\textsuperscript{*} & MoLe-CogAct & CAPER \\
\midrule
Pick cup      & 40.0 & 84.0 & 86.0 & 85.2 & 48.0 \\
Push button   & 48.0 & 82.4 & 84.0 & 88.0 & 80.0 \\
Take umbrella & 20.0 & 28.0 & 32.0 & 36.0 & 42.4 \\
Put knife     & 16.0 &  8.0 &  10.4 & 26.4 & 73.2 \\
Put money     & 44.0 & 22.4 & 24.0 & 30.0 & 46.4 \\
\bottomrule
\end{tabularx}
\textsuperscript{*}{\small{Re-implemented following the paper, as official code is unavailable.}}
\caption{Success rate (\%) of different methods on RLBench.}
\label{tab:Success rate of RLBench experiments}
\end{table*}

From the experimental results, we observe the critical role of visual information and CoT reasoning structure within the CAPER framework. 
Moreover, each module in CAPER proves to be indispensable. 
The success rate of CAPER in scientific experiment tasks meets the practical requirements of such experiments.

\subsection{Experimental results on the scientific experiment dataset (real-world)}



We design a sim-to-real setup where real-world objects differ in shape from simulation but share the same sizes and spatial layouts.
Object poses are obtained using a D415 depth camera with YOLOv8 and mapped to the simulation frame.

For real-world evaluation, we conduct 40 trials each on the tasks {Pick\_place}, {Pour}, and {Stir}, 20 trials on the extra-long-horizon tasks {Mixing} and {Crystallize}, and 10 trials on the real-only tasks {Weigh} and {Shake}.
CAPER achieves success rates of 75.0\%, 62.5\%, 70.0\%, 50.0\%, 25.0\%, 70.0\%, and 70.0\% on these tasks, respectively.

\begin{figure}[t]
\centering
\includegraphics[width=\linewidth]{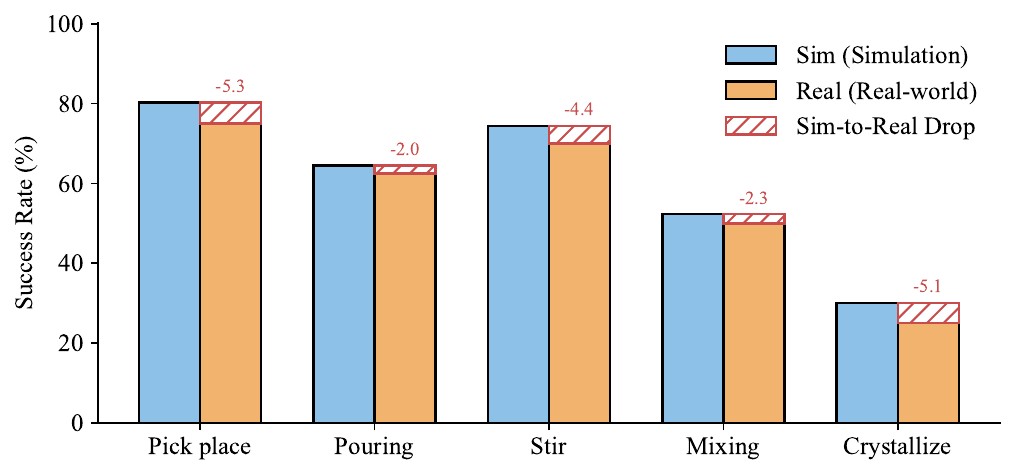}
\caption{Sim-to-real performance comparison \& drop analysis.}
\label{fig: sim2real}
\end{figure}

As shown in Fig.~\ref{fig: sim2real}, the performance gap between simulation and real-world execution remains moderate.
Importantly, the observed drop is mainly caused by inaccuracies in object detection and depth estimation, rather than limitations of task reasoning or execution.
Such errors are confined to the perception module and can be mitigated by improving sensing accuracy, without retraining the task-level planner or the low-level controller.
Notably, the {Weigh} and {Shake} tasks are executed purely in real-world settings, without any corresponding simulation environment.
These tasks are performed using the same action primitives and execution policies learned in simulation, without additional RL training.
This result indicates that CAPER supports direct deployment in real-world scenarios where simulation is unavailable, by decoupling procedural structure and execution from environment-specific perception.
Further analysis is provided in Appendix. 

\subsection{Experimental results on the public dataset}
We further evaluate CAPER on RLBench~\citep{james2020rlbench}, a widely used robotic manipulation benchmark.
To differentiate CAPER from CoT-VLA, and owing that the code of CoT-VLA is not publicly available, we reproduced the entire pipeline based on the VILA-U model \citep{DBLP:conf/iclr/WuZCTLFZXYY0025} described in the CoT-VLA paper.
To ensure a fair comparison, all models are trained on the same task-specific training data without pretraining on real-world demonstrations or action-free videos, which may negatively impact performance in simulation.



We select 5 long-horizon tasks that closely align with the typical procedures of scientific experiments in RLBench to validate the performance of CAPER, and compare the success rate with PerAct \citep{shridhar2023perceiver} , OpenVLA \citep{DBLP:conf/corl/KimPKXB0RFSVKBT24}, CoT-VLA \citep{zhao2025cot} and MoLe-CogAct \citep{DBLP:journals/corr/abs-2503-20384} methods.
We evaluate the success rates of CAPER and other methods using the benchmark we developed, presenting the average success rate across 3 evaluation rounds, each with 25 trials.
As shown in Table~\ref{tab:Success rate of RLBench experiments}, CAPER achieves the highest success rates on long-horizon tasks, demonstrating stronger robustness and adaptability to procedural task variations.
We note that performance on the {pick cup} task is lower than on other tasks, primarily due to a conservative grasping constraint inherited from the scientific experiment setting, where the RL controller is restricted to grasping only one side of the cup wall for safety.

\subsection{Model efficiency and runtime}

Most components of CAPER do not require training.
All LLM and VLM modules are used off-the-shelf, while only the multimodal predictor and the RL controller are trained.
Among them, the multimodal predictor is the most computationally expensive component, whereas the RL policy can be trained on a single consumer-grade GPU (e.g., RTX 4060).
Inference latency of each module is reported in Table~\ref{tab:inference_latency}.

To assess whether CAPER can be executed more efficiently, we additionally evaluate a reduced resolution of $128 \times 128$ (inference cost 978ms).
We observe comparable performance under the reduced resolution, indicating that CAPER can be accelerated without significant loss of effectiveness.
Details are provided in Appendix. 

\begin{table}[tbp]
\centering
\setlength{\tabcolsep}{15pt}
\renewcommand{\arraystretch}{1.1} 
\begin{tabular}{lcc}
\hline
Component    & \multicolumn{1}{l}{Latency (ms)} & \multicolumn{1}{l}{RunsOnline?} \\ \hline
LLM          & 3024                            & No                              \\
MP           & 5714                            & Optional                        \\
VLM          & 3441                            & Optional                        \\
RL           & 15                              & Yes                             \\ \hline
Online total & 15                              & -                               \\ \hline
\end{tabular}
\caption{Per-module inference latency and online/offline execution.}
\label{tab:inference_latency}
\end{table}
\section{Conclusion}


We presented CAPER, a framework for robotic assistance in scientific R\&D laboratories that emphasizes explicit procedural structure and responsibility separation in long-horizon manipulation. By structuring task-level reasoning, perceptual grounding, and low-level control into components with well-defined roles, CAPER addresses a key mismatch between end-to-end VLAs and the requirements of protocol-sensitive experimentation.
Experiments on a scientific workflow benchmark and a public long-horizon manipulation dataset show that CAPER improves robustness and procedural correctness, particularly in low-data and long-horizon settings. These results suggest that explicitly constraining where reasoning occurs can lead to more reliable and controllable robotic execution in scientific workflows.

\bibliographystyle{named}
\bibliography{ijcai26}

\clearpage
\appendix









\appendix

\section*{Appendix}
\section{LLM Usage Statement}
Large Language Models (LLMs) are used to assist in language polishing and improving the readability of the manuscript.

\section{Detailed Description of the Self-verification Operations}
\label{app:self-ver}

\paragraph{Subgoal Validation.}  The LLM is prompted to check whether each subtask aligns with the overall task objective through natural-language self-reflection and identify potential contradictions, redundancies, or omissions.
\paragraph{Constraint Consistency Check.} For each subtask, we guide the model to ensure each subtask satisfies key physical, temporal, and causal constraints, detecting invalid ordering (e.g., measuring before synthesis) or mismatches (e.g., heating before reagent addition).
\paragraph{Correction and Refinement.}  When inconsistencies are identified, the model is prompted to revise and revalidate subtasks until the sequence is logically coherent and operationally feasible.

\section{Discussion of Why and When Decoupling is Valid}
\label{app:decouple}

In this section, we provide a brief theoretical analysis of the conditions under which the decoupling between the task-level symbolic planner and downstream execution is valid. 
We focus on the subtask sequence generated by the task-level planner and analyze its contribution to modularity, while noting that downstream action primitives and RL policies do not affect this argument.

\paragraph{Preliminaries.}
Let $\mathcal{G}$ denote a high-level task goal, $\mathcal{S} = (s_1, s_2, \dots, s_T)$ the symbolic subtask sequence generated by the task-level planner, and $\pi$ a low-level execution policy acting on continuous observations $o_{1:T}$. 
Task success is represented by the event $\mathcal{E} = \{\text{goal achieved}\}$.

\begin{lemma}[Task-symbolic sufficiency]
If the execution policy $\pi$ depends only on observations $o_{1:T}$ given $\mathcal{S}$, i.e.,
\[
P(\pi \mid \mathcal{S}, \mathcal{G}, o_{1:T}) = P(\pi \mid \mathcal{S}, o_{1:T}),
\]
then the goal $\mathcal{G}$ influences execution only through the symbolic plan $\mathcal{S}$. 
\end{lemma}

\begin{lemma}[Conditional independence]
If symbolic decompositions and observations are conditionally independent given the goal,
\[
P(\mathcal{S}, o_{1:T} \mid \mathcal{G}) = P(\mathcal{S}\mid \mathcal{G}) \, P(o_{1:T}\mid \mathcal{G}),
\]
then $\mathcal{S}$ carries no redundant information about $o_{1:T}$ beyond $\mathcal{G}$. 
\end{lemma}

\begin{proposition}[Factorization of success probability]
Under Lemmas 1 and 2, the probability of success can be factorized as
\begin{align}
    \mathcal{E} &:=
    \max P(\pi \mid \mathcal{G}, o_{1:T}), \\
    P(\pi \mid \mathcal{G}, o_{1:T}) &= \sum_{\mathcal{S}} P(\pi, \mathcal{S} \mid \mathcal{G}, o_{1:T}) \\
    &= \sum_{\mathcal{S}} P(\pi\mid \mathcal{S}, o_{1:T}) \, P(\mathcal{S}\mid \mathcal{G}).
\end{align}
Hence the contribution of reasoning (via $P(\mathcal{S}\mid \mathcal{G})$) is \emph{decoupled} from perception and control (via $P(\mathcal{E}\mid \mathcal{S}, o_{1:T})$).
\end{proposition}

This shows that improvements in task-level reasoning and low-level execution independently contribute to overall success, consistent with modular planning frameworks \citep{DBLP:journals/ai/KaelblingLC98}.

\paragraph{When decoupling is valid.}
Decoupling holds if:
\begin{itemize}
    \item Each subtask $s_i$ contains sufficient symbolic information for execution without direct dependence on $\mathcal{G}$;
    \item Observations and subtasks are approximately conditionally independent given $\mathcal{G}$;
    \item Subtasks capture the causal structure of the task relevant for achieving the goal.
\end{itemize}

For example, for a task like “mix reagents A and B”, the task-level planner may output subtasks such as \textit{add reagent A} and \textit{add reagent B}. 
Each subtask carries enough symbolic information for downstream execution, the low-level observations for executing these actions are largely independent of the high-level goal, and the sequence preserves the necessary causal order. 
Compared to a fully end-to-end approach that maps from goal and observations directly to actions, decoupling reduces the complexity and naturally aligns with the modular structure of the task.

Overall, while the conditions are not guaranteed in all settings, they are relatively easy to satisfy in well-structured tasks, which explains why decoupling is a practical and effective design choice.


\paragraph{When decoupling may fail.}
In practice, these assumptions may be violated in cases such as:
\begin{itemize}
    \item If $\mathcal{S}$ omits crucial physical requirements (e.g., exact forces or contact stability), then $P(\mathcal{E}\mid \mathcal{S}, o_{1:T})$ still implicitly depends on $\mathcal{G}$.
    \item If $\mathcal{S}$ is entangled with perception (e.g., symbolic labels directly derived from raw visual embeddings), then independence may fail.
    \item If success depends on hidden histories not captured in $\mathcal{S}$, factorization does not hold.
\end{itemize}

\paragraph{Fallback as constrained refinement.}
Execution failures can be modeled as the discovery of additional constraints $C$ on symbolic decompositions. 
Formally,
\[
\mathcal{S}^{(k+1)} \sim P(\mathcal{S}^{(k)} \mid \mathcal{G}, C),
\]
where $C$ is induced from observed failure conditions (e.g., contact instability). 
If the constraint set $\mathcal{C}$ is finite and each refinement step reduces the feasible set of $\mathcal{S}$, then iterative fallback converges in at most $|\mathcal{C}|$ iterations: either a feasible $\mathcal{S}^\star$ is found, or infeasibility is proven. 
This parallels refinement strategies in symbolic AI.

\paragraph{Limitations and outlook.}
Our current system does not implement fallback at this stage, i.e., the planner runs only once. 
Nonetheless, the above analysis suggests a principled extension: failed executions could be recycled as symbolic constraints to refine subtask decomposition. 
Future work may explore this integration, which would tighten the link between symbolic reasoning and embodied feedback while preserving modularity.

\section{Training Details}
We implemented our multimodal predictor using PyTorch. 
The backbone of the diffusion model is U-Net. 
The multimodal predictor is conducted with the following configuration:
\begin{itemize}
    \item Number of diffusion steps: 100
    \item Sampling steps: 100
    \item loss: L2
    \item SNR-based loss weighting: True
    \item Image size: $(256,256)$
\end{itemize}
We adopt OpenAI’s CLIP as the text encoder, with its parameters kept frozen throughout training to preserve the pretrained semantic representations.
The multimodal predictor is trained on an NVIDIA RTX 4090 GPU, and the model checkpoints are saved every 2,500 iterations.
The training is conducted with the following configuration:
\begin{itemize}
    \item Learning rate: 1e-5
    \item Batch size: 1 
    \item Image sequence length: 8
    \item Total training steps: 60,000
    \item Precision: Mixed precision (fp16) 
\end{itemize}
The training process of multimodal predictor takes approximately three days on our collected dataset.


 \section{Results of Different Large Models}
Since only relying on large language models to evaluate outputs can be inaccurate, which often leading to redundant content.
Thus we assessed the performance of different LLMs and VLMs on our defined tasks through human evaluation.
For each model, we prompt the large model to generate 25 task decompositions in each round, and compute the corresponding success rate based on expert evaluations. 
The final success rate reported in the table is the average of 3 rounds.
We use LLaMA3.1 as the LLM and GPT‑4o as the VLM in our framework.

\begin{table}[htbp]
\caption{Success rate (\%) of different LLMs}
\centering
\setlength{\tabcolsep}{3pt} 
\begin{tabularx}{\linewidth}{lccccc}
\toprule
Model & Pick\_place & Pour & Stir & Mix & Crystallize \\
\midrule
Qwen3              & 45.2  & 36.0 & 100.0   & 88.0 & 56.0 \\
Deepseek-V3        & 98.4  & 44.0 & 100.0   & 33.2 & 90.4 \\
Llama3.1 (ours)    & 100.0   & 74.4 & 100.0   & 73.2 & 76.0 \\
\bottomrule
\end{tabularx}

\label{tab:Success rate of different large language models}
\end{table}

\begin{table}[htbp]
\caption{Success rate (\%) of different VLMs}
\centering
\setlength{\tabcolsep}{3pt} 
\renewcommand{\arraystretch}{1.2} 
\begin{tabularx}{\linewidth}{lccccc}
\toprule
Model & Pick\_place & Pour & Stir & Mix & Crystallize \\
\midrule
Qwen2.5          & 77.2  & 76.0  & 76.0  & 73.2  & 72.0 \\
Llama3.2         & 82.4  & 80.0  & 82.4  & 78.4  & 76.0 \\
GPT-4o (ours)    & 100.0   & 100.0   & 100.0   & 98.4  & 98.4 \\
\bottomrule
\end{tabularx}
\label{tab:Success rate of different VLMs}
\end{table}

\begin{table*}[htbp]
\caption{Success rate (\%) of CoT reasoning and its ablations (w/o Step1--4)}
\centering
\setlength{\tabcolsep}{16pt} 
\renewcommand{\arraystretch}{1.2} 
\begin{tabularx}{\linewidth}{lccccc}
\toprule
Task & CoT & CoT w/o Step1 & CoT w/o Step2 & CoT w/o Step3 & CoT w/o Step4 \\
\midrule
Pick\_place   & 100.0 & 44.0 & 37.2 & 46.4 & 41.2 \\
Pour          & 74.4  & 13.2 & 22.4 & 25.2 & 22.4 \\
Stir          & 100.0 & 57.2 & 29.2 & 78.4 & 61.2 \\
Mix        & 73.2  & 40.0 & 25.2 & 66.4 & 13.2 \\
Crystallize   & 76.0  & 70.4 & 49.2 & 54.4 & 30.4 \\
\bottomrule
\end{tabularx}
\label{tab:Success rate of  CoT reasoning}
\end{table*}

\section{Task Details}


Our collected dataset includes five task types defined in simulation: {Pick\_place}, {Pour}, {Stir}, {Mix}, and {Crystallize}, as well as two real-world–only tasks: {Weigh} and {Shake}.
To support these tasks, we define four action primitives: \texttt{move}, \texttt{grasp}, \texttt{pour}, and \texttt{stir}, which capture core operations commonly involved in scientific experimental procedures.
All visual data are collected at a resolution of $512 \times 512$.


\begin{itemize}
    \item {\textbf{Pick\_place}:} The {pick\_place} task includes 3 types of object manipulation tasks: moving the cuboid to the beaker, moving the cuboid to the glass garden, and moving the cylinder to the beaker.  
    In these tasks, we use cuboids and cylinders to represent objects with different shapes and graspable properties. 
    Fig.~\ref{fig:pick_place} illustrates example images from the {pick\_place} task.
    \item {\textbf{Pour}:} Considering the specific procedural requirements of certain scientific experiments, we collected a visual demonstration set for the {pour} task.  We use cups as the manipulable object to represent the action of pouring.  
    The execution of the {pour} operation requires prior completion of the \texttt{move} and \texttt{grasp} operations from the {pick\_place} task.
    \item {\textbf{Stir}:} The {stir} task is designed to perform liquid stirring operations.
    A thin rectangular block is used as a graspable stirrer to ensure the successful execution of the {stir} operation.  
    Similar to {pour}, the {stir} operation also requires prior completion of the \texttt{move} and \texttt{grasp} operations.
    \item {\textbf{Mix}:} {Mix} is a longer-horizon task. 
    Considering the frequent use of reactant mixing in scientific experiments, we incorporate this task into our setup.  
    This long-horizon task can be decomposed and accomplished through a combination of the three previously defined action primitives.
    \item {\textbf{Crystallize}:} The {crystallize} task has a similar horizon length to the {mix} task, both being more complex than the former 3 tasks.  
    This task can also be decomposed into a sequence of our designed action primitives.  
    Unlike the \texttt{mix} task, however, the {crystallize} task requires more frequent use of {pick\_place} operations rather than {pour}.
    \item {\textbf{Weigh}:} The {weigh} task is a real-world–only task that measures the weight of a container using a laboratory balance.  
    Due to the difficulty of accurately modeling specific instruments in simulation, this task is executed directly in the real environment and relies on visual feedback to complete the weighing procedure.
    \item {\textbf{Shake}:} The {shake} task is another real-world–only task, where a container is placed onto a laboratory shaker to perform mixing through shaking.  
    Similar to {weigh}, this task is executed without a corresponding simulation environment and is realized through a sequence of predefined action primitives.
\end{itemize}

\section{CoT Prompt design}

Our CoT reasoning of the LLM is designed in 4 steps. Each step guides the LLM to think through how to properly decompose the long-horizon task and evaluate whether the resulting short-horizon subtasks are sufficient to accomplish the original objective. 
This decomposition process does not require visual input.

\begin{figure*}[htbp]
\centering
\begin{subfigure}[b]{0.72\linewidth}
    \centering
    \includegraphics[width=\linewidth]{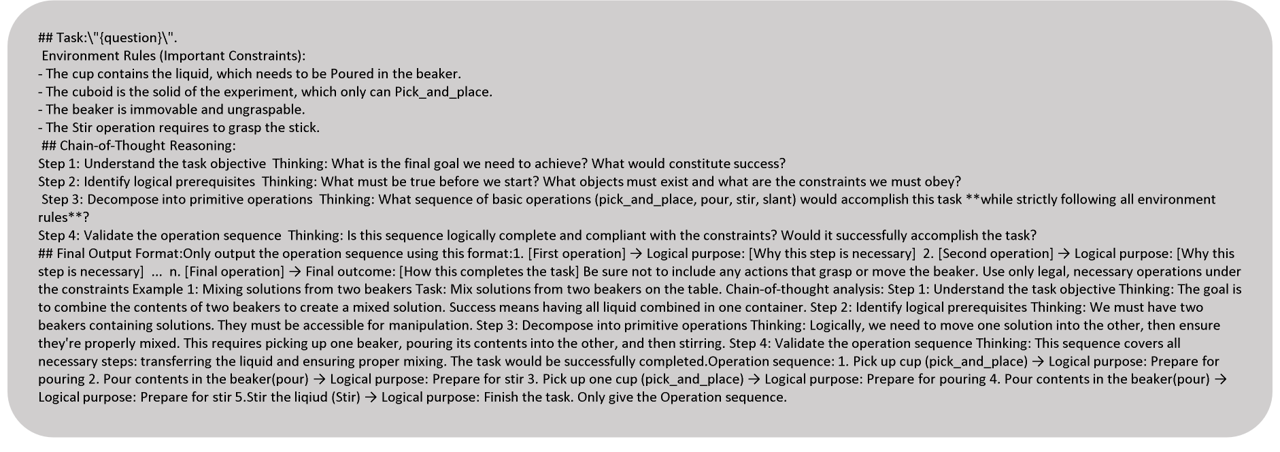}
    \caption{Design of the CoT prompt}
    \label{fig:s_cot}
\end{subfigure}


\begin{subfigure}[b]{0.72\linewidth}
    \centering
    \includegraphics[width=\linewidth]{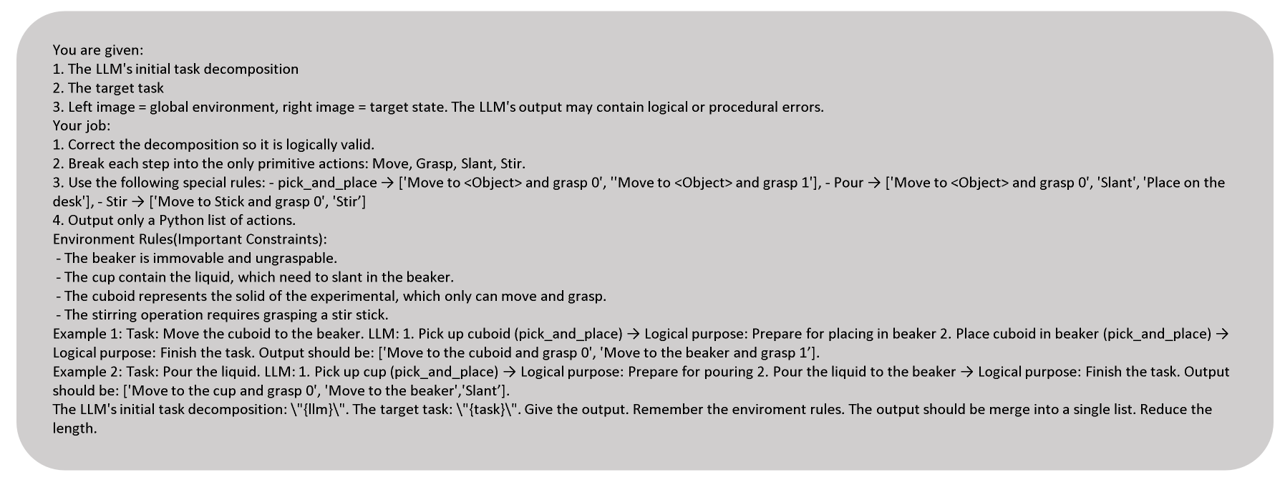}
    \caption{Design of the VLM prompt}
    \label{fig:s_mlm}
\end{subfigure}

\caption{Prompt designs for different reasoning strategies: (a) CoT and (b) VLM.}
\label{fig:prompt_designs}
\end{figure*}

Instead, the model relies on commonsense reasoning, following the structured CoT prompts to complete the task breakdown. 
The complete reasoning chain is illustrated in Fig. \ref{fig:s_cot}.

We also conducted distillation experiments on each of the four stages to demonstrate that they are essential components in the task decomposition process.
The success rate is evaluated in the same manner as the section of ``Results of Different Large Models''.

We evaluated the task decomposition success rate of CoT reasoning when each module is ablated. 
The result is shown in Table \ref{tab:Success rate of  CoT reasoning}.

Meanwhile, we defined two common types of errors for analysis: Step Redundant Error (E1) and Step Logical  Error (E2), allowing for a more detailed evaluation of decomposition quality.
We further record the occurrences of two defined error types, which are shown in Table \ref{tab:error_rates}.
We use the crystallization cultivation experiment as an example to illustrate the two error types: E1 and E2.



\begin{figure}[h]
\centering
\begin{subfigure}[b]{1.0\linewidth}
    \includegraphics[width=\linewidth]{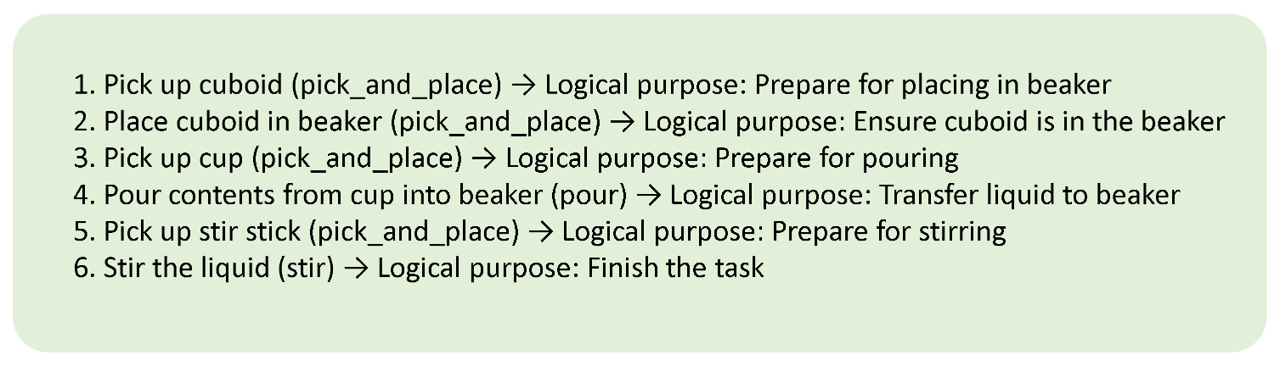}
    \caption{Correct CoT output}
    \label{fig:cot_correct}
\end{subfigure}


\begin{subfigure}[b]{1.0\linewidth}
    \includegraphics[width=\linewidth]{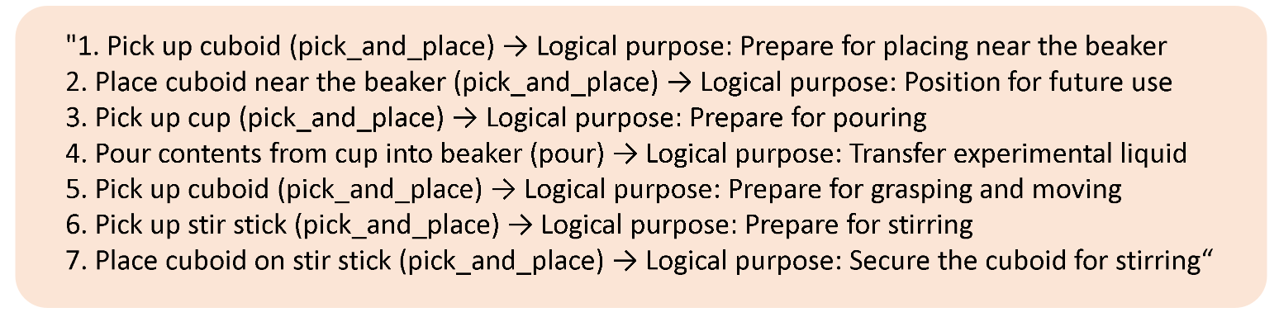}
    \caption{E1 error example: redundant step}
    \label{fig:E1}
\end{subfigure}


\begin{subfigure}[b]{1.0\linewidth}
    \includegraphics[width=\linewidth]{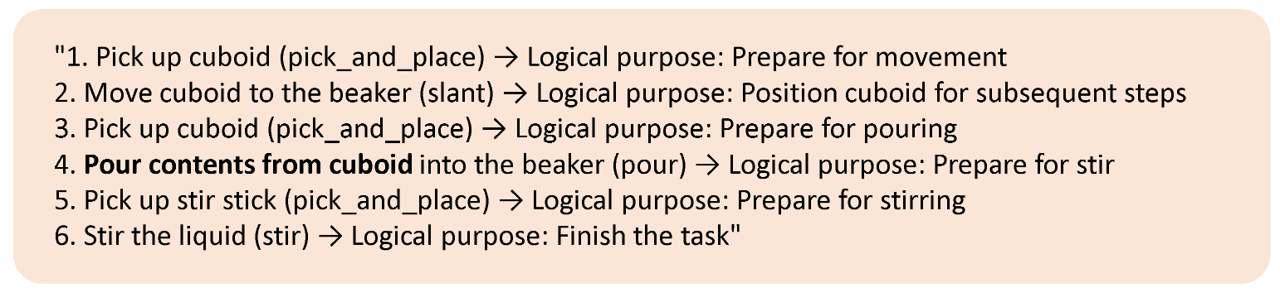}
    \caption{E2 error example: incorrect logic}
    \label{fig:E2}
\end{subfigure}

\caption{Examples of CoT module outputs: (a) correct, (b) E1 error, (c) E2 error.}
\label{fig:cot_outputs}
\end{figure}

The correct CoT module output for the crystallization cultivation experiment is shown in 
Fig.~\ref{fig:cot_correct}.
We also provide example output images illustrating {E1} and {E2} errors, which are shown in 
Figs. \ref{fig:E1} and \ref{fig:E2}.
In the {E1} example, the model repeatedly outputs the instruction ``pick up cuboid''.
Although the overall sequence is structurally complete, the redundancy in a single step leads to its classification as an {E1} error.
In the {E2} example, the model outputs ``pour contents from cuboid'' when there is no object inside the cuboid.
This results in a logically incorrect step, thus leading to its classification as an {E2} error.










\section{Efficiency Validation}
\label{app:efficiency}

This appendix provides supplementary validation experiments on the efficiency of CAPER under reduced multimodal predictor resolution.
While all main experiments are conducted with an MP input resolution of $256 \times 256$, we retrain the MP at a lower resolution to evaluate the feasibility of accelerating execution (Table \ref{tab:efficiency}).

We retrain the multimodal predictor with an input resolution reduced from $256 \times 256$ to $128 \times 128$, while keeping all other components unchanged.
In particular, the task-level planner, VLM, and low-level RL controller are directly reused without retraining.
We evaluate the retrained MP on the long-horizon tasks {Mixing} and {Crystallize}.
Results show only minor differences in success rates compared to the original resolution, indicating that the procedural structure and execution policies of CAPER are robust to changes in visual prediction fidelity.

Under the reduced resolution setting, we further explore an optional strategy that provides newly observed images to the MP during execution to reduce error accumulation.
This strategy yields a modest improvement in success rates on long-horizon tasks.
However, due to the inference latency of the MP, the overall gain remains limited.
As a result, online visual updates are not enabled in the main experiments.

\begin{table*}[htbp]
\caption{Error rates (\%) of E1 and E2 under CoT reasoning and its ablations (w/o Step1--4)}
\centering
\setlength{\tabcolsep}{12pt} 
\renewcommand{\arraystretch}{1.2} 
\begin{tabularx}{\linewidth}{l *{10}{r}}
\toprule
\multirow{2}{*}{Task} & \multicolumn{2}{c}{CoT} & \multicolumn{2}{c}{CoT w/o Step1} & \multicolumn{2}{c}{CoT w/o Step2} & \multicolumn{2}{c}{CoT w/o Step3} & \multicolumn{2}{c}{CoT w/o Step4} \\
                       & E1 & E2 & E1 & E2 & E1 & E2 & E1 & E2 & E1 & E2 \\ 
\midrule
Pick\_place   & 0.0  & 0.0  & 54.4 & 26.4 & 62.4 & 10.4 & 42.4 & 30.4 & 58.4 & 20.0 \\
Pour          & 2.6  & 25.6 & 86.4 & 18.4 & 76.0 & 9.2  & 70.4 & 34.4 & 77.2 & 30.4 \\
Stir          & 0.0  & 0.0  & 42.4 & 25.2 & 66.4 & 34.4 & 14.4 & 6.4  & 39.8 & 0.0  \\
Mix        & 8.0  & 26.8 & 56.0 & 46.4 & 10.4  & 76.0 & 9.2  & 33.2 & 0.0  & 86.8 \\
Crystallize   & 10.0  & 14.0 & 0.0  & 29.6 & 18.4 & 38.4 & 6.4  & 34.4 & 20.0 & 69.2 \\
\bottomrule
\end{tabularx}
\label{tab:error_rates}
\end{table*}


\begin{table}[t]
\centering
\small
\setlength{\tabcolsep}{3pt}
\begin{tabular}{lcc}
\toprule
Setting & {Mixing} & {Crystallize} \\
\midrule
MP ($256 \times 256$)         & 52.3 & 30.1 \\
MP ($128 \times 128$)         & 51.6 & 29.8 \\
MP ($128 \times 128$) (online) & 55.2 & 30.6 \\
\bottomrule
\end{tabular}
\caption{Validation of reduced MP resolution and online visual updates on long-horizon tasks.}
\label{tab:efficiency}
\end{table}

\section{VLM Prompt design}

The prompt design for the VLM is primarily aimed at mapping the subtask instructions generated by the large language model to our predefined three action primitives, in order to guide the robot's motion accordingly. The detailed prompt structure is illustrated in Fig. \ref{fig:s_mlm}.

\section{Experiment in real-world scenario}
\label{app:realworld}

\begin{figure*}[htbp]
\centering
\includegraphics[height=0.27\textheight]{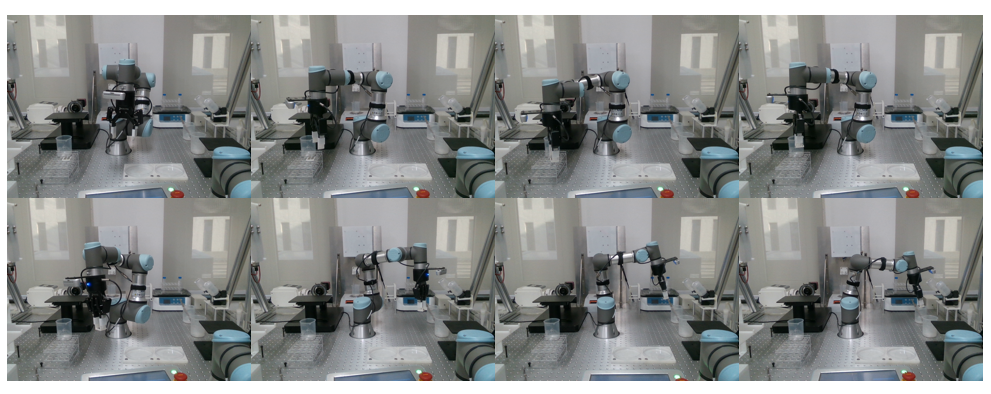}
\caption{``Pick\_place'' task in real-world scenario.}
\label{fig:real-world pick_place}
\end{figure*}

\begin{figure*}[htbp]
\centering
\includegraphics[height=0.27\textheight]{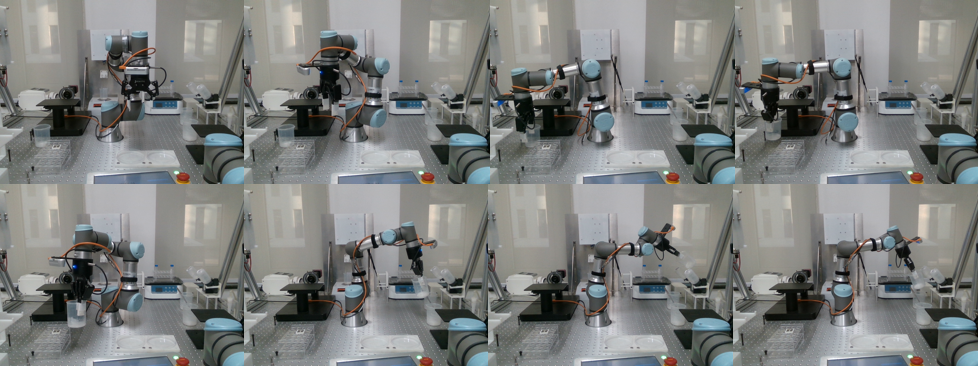} \\[-0.5ex]
\includegraphics[height=0.27\textheight]{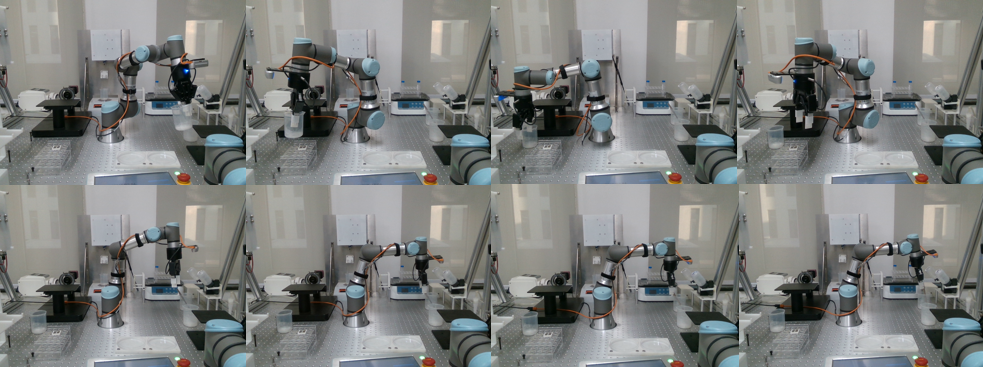}
\caption{``Crystallize'' task in real-world scenario.}
\label{fig:real-world crystallize}
\end{figure*}

\begin{figure*}[htbp]
\centering
\includegraphics[height=0.27\textheight]{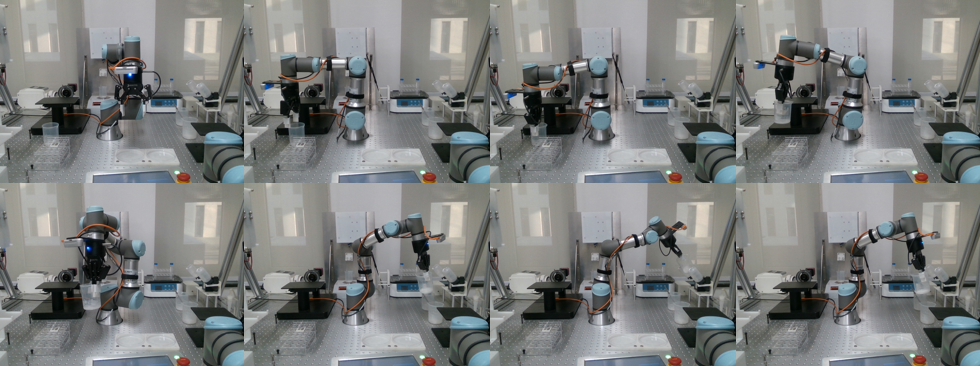} \\[-0.5ex]
\includegraphics[height=0.27\textheight]{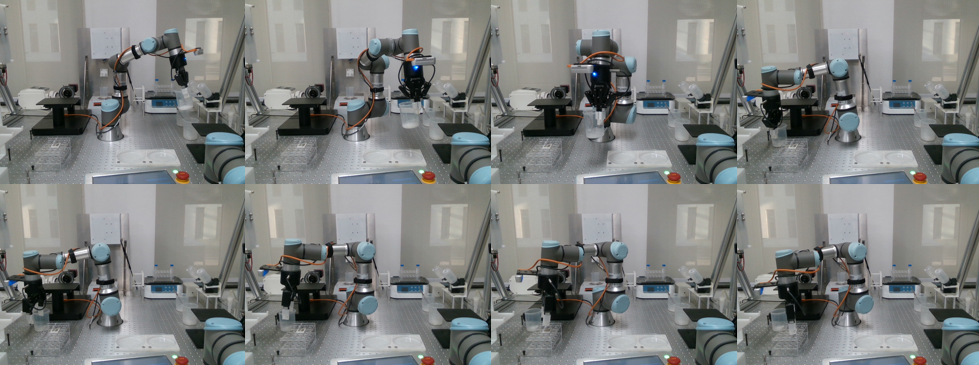} \\[-0.5ex]
\includegraphics[height=0.27\textheight]{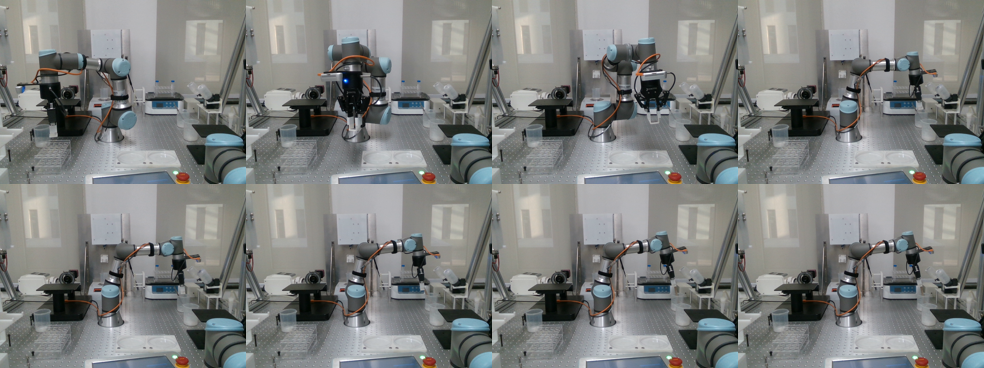}
\caption{``Mix'' task in real-world scenario.}
\label{fig:real-world mixing}
\end{figure*}

\begin{figure*}[htbp]
\centering
\includegraphics[height=0.27\textheight]{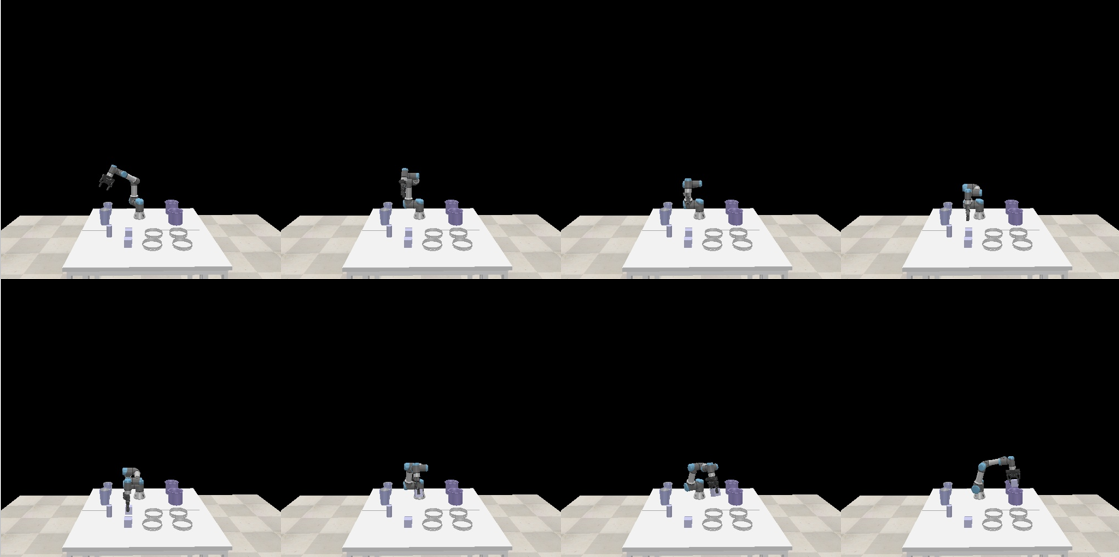} \\[-0.5ex]
\includegraphics[height=0.27\textheight]{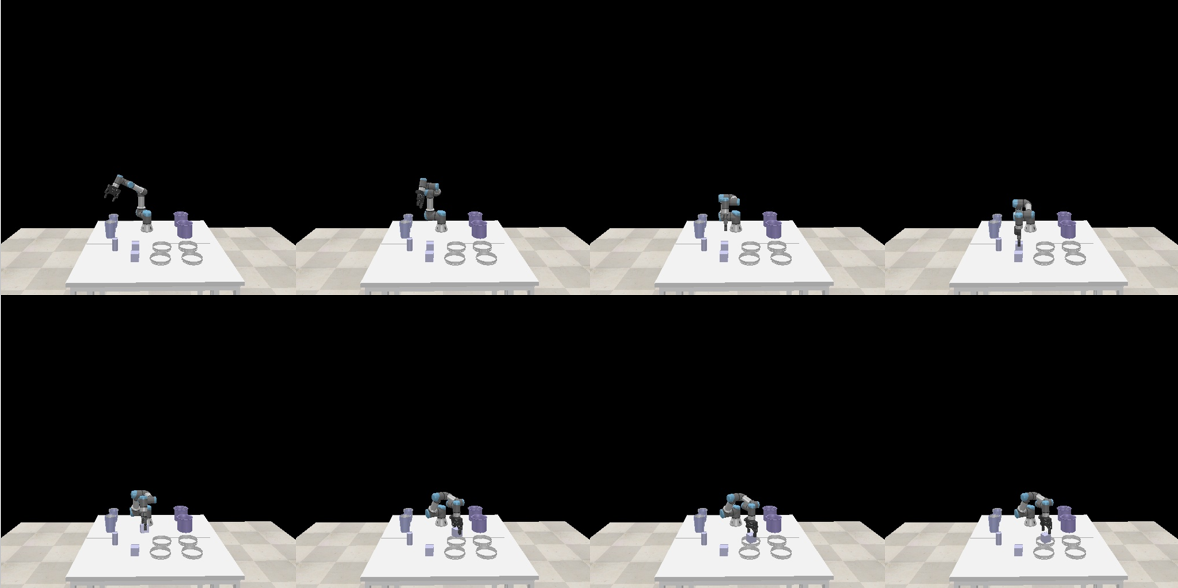} \\[-0.5ex]
\includegraphics[height=0.27\textheight]{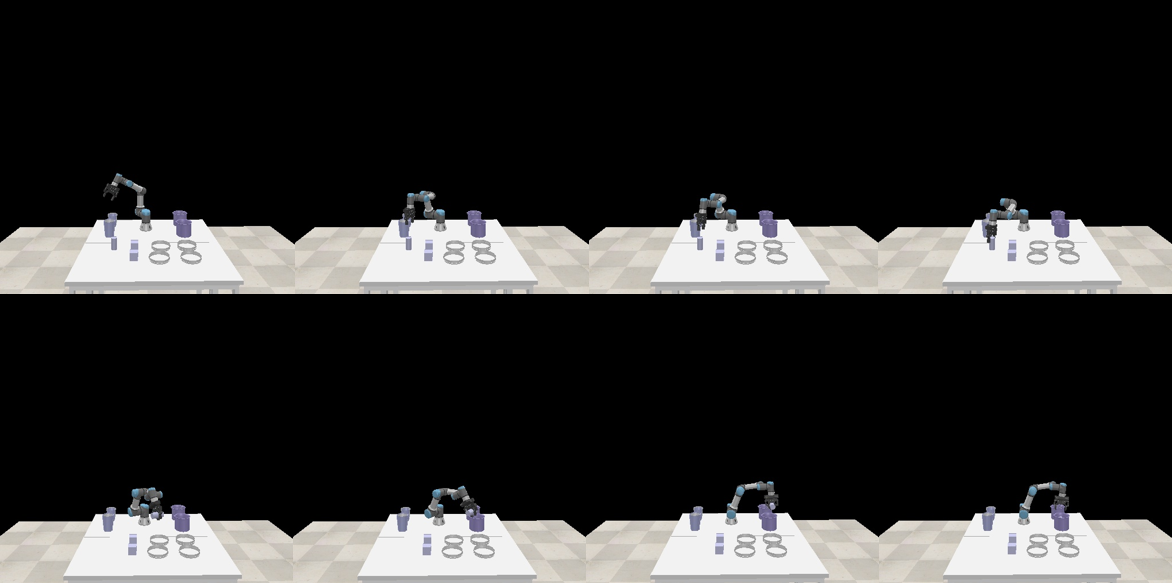}
\caption{Visualization of the ``pick\_place'' task.}
\label{fig:pick_place}
\end{figure*}

\begin{figure*}[htbp]
\centering{\includegraphics[width=0.9\linewidth]{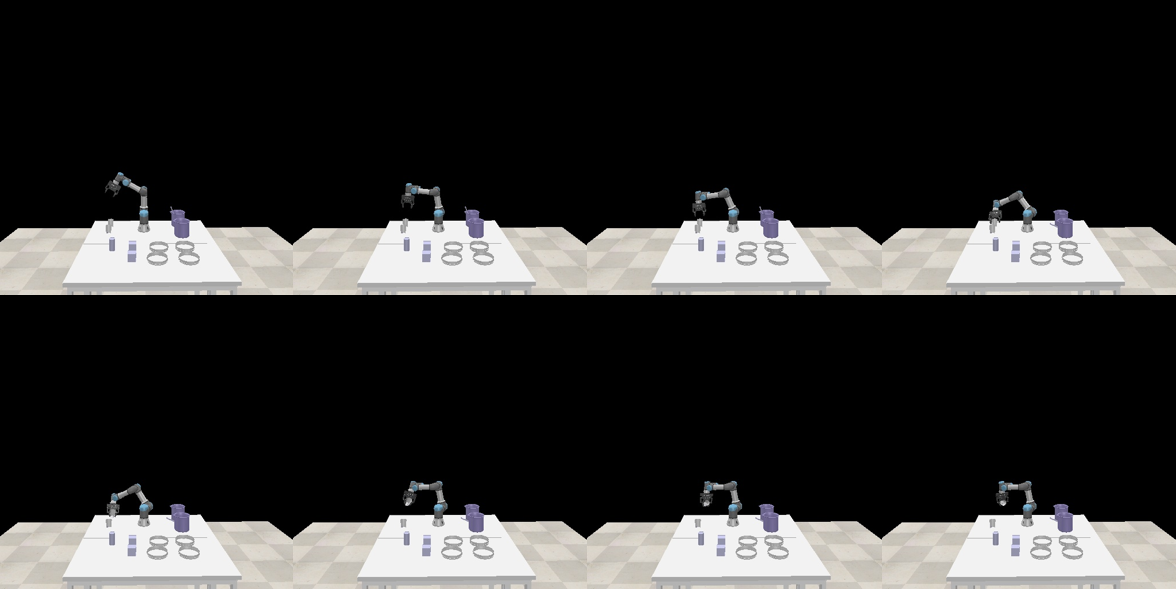}}
\caption{Visualization of the ``pour'' task.}
\label{fig:Pour}
\end{figure*}

\begin{figure*}[htbp]
\centering{\includegraphics[width=0.9\linewidth]{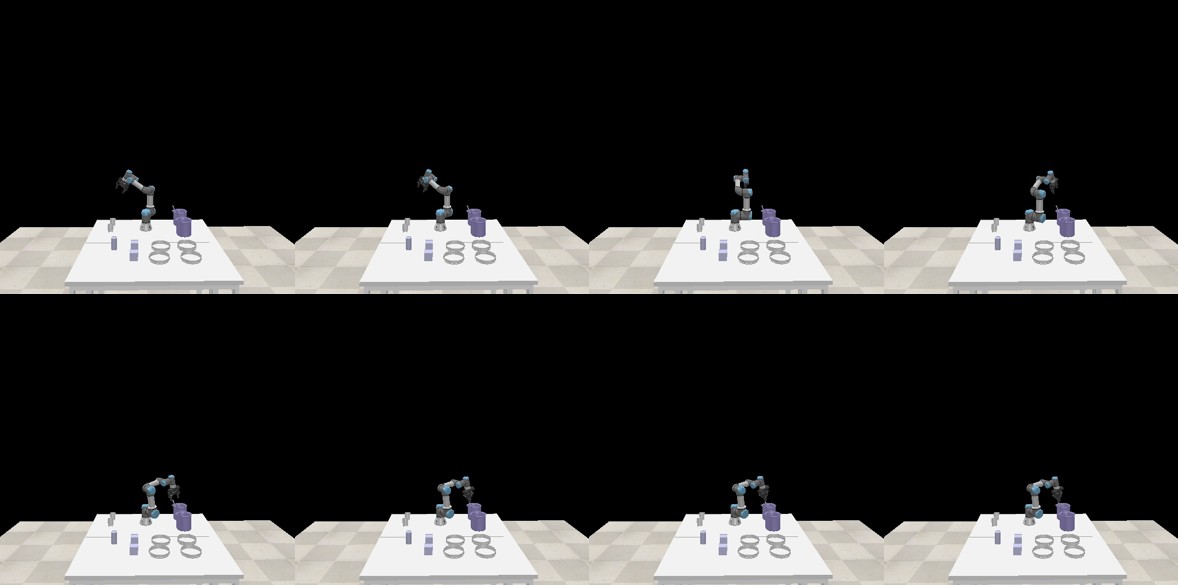}}
\caption{Visualization of the ``stir'' task.}
\label{fig:Stir}
\end{figure*}

\begin{figure*}[htbp]
\centering{\includegraphics[width=0.9\linewidth]{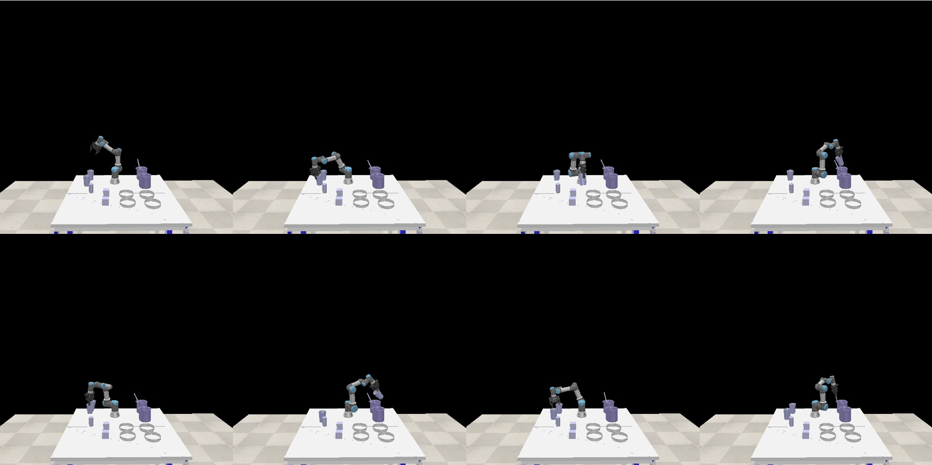}}
\caption{Visualization of the ``mix'' task.}
\label{fig:Mix}
\end{figure*}

\begin{figure*}[htbp]
\centering{\includegraphics[width=0.9\linewidth]{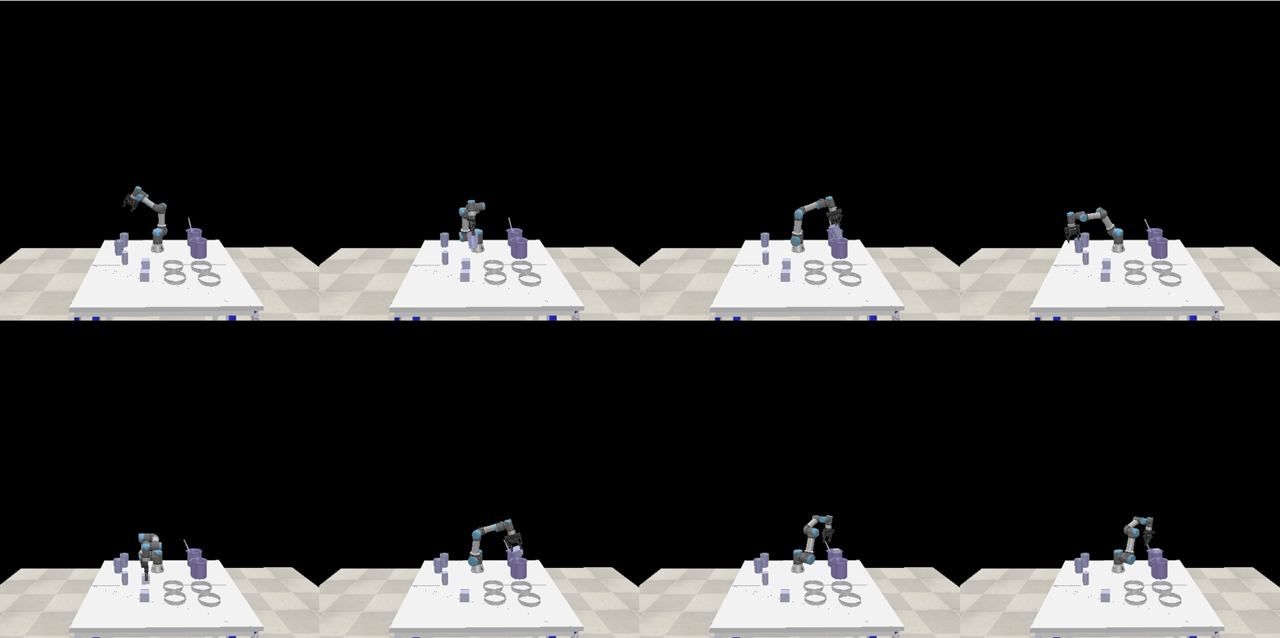}}
\caption{Visualization of the ``crystalize'' task.}
\label{fig:Crystallize}
\end{figure*}

\begin{figure*}[htbp]
\centering
\includegraphics[width=0.7\textheight]{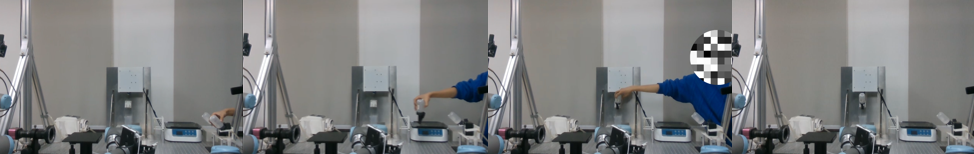}
\caption{Human demonstration of the ``weigh'' task in real-world scenario.}
\label{fig:real-world human_weigh}
\end{figure*}

\begin{figure*}[htbp]
\centering
\includegraphics[height=0.27\textheight]{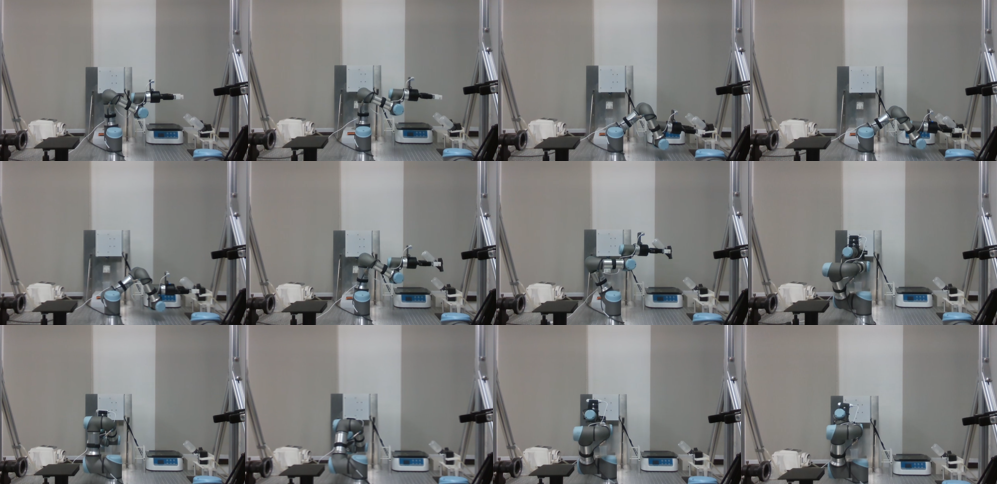}
\caption{``Weigh'' task in real-world scenario.}
\label{fig:real-world weigh}
\end{figure*}

\begin{figure*}[htbp]
\centering
\includegraphics[height=0.27\textheight]{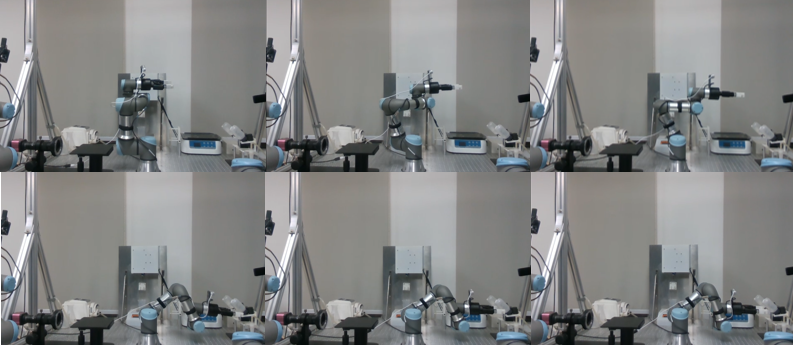}
\caption{ Failure case of the ``weigh'' task in real-world scenario.}
\label{fig:real-world weigh_mistake}
\end{figure*}

\begin{figure*}[htbp]
\centering
\includegraphics[width=0.7\textheight]{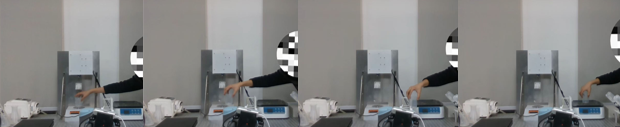}
\caption{Human demonstration of the ``shake'' task in real-world scenario.}
\label{fig:real-world human_shake}
\end{figure*}

\begin{figure*}[htbp]
\centering
\includegraphics[width=0.7\textheight]{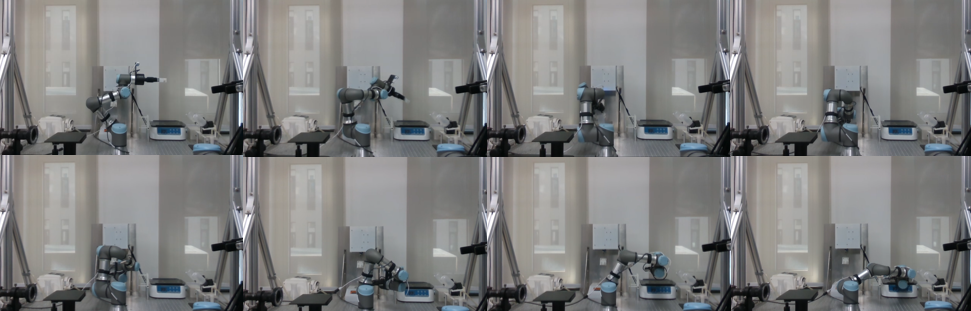}
\caption{``Shake'' task in real-world scenario.}
\label{fig:real-world shake}
\end{figure*}

\begin{figure*}[htbp]
\centering
\includegraphics[width=0.7\textheight]{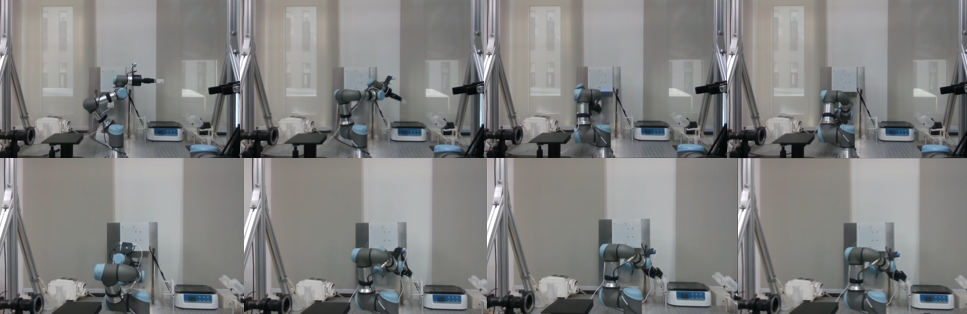}
\caption{ Failure case of the ``shake'' task in real-world scenario.}
\label{fig:real-world shake_mistake}
\end{figure*}

We replicate the experimental setup from the simulation environment in a real-world setting and evaluate the success rate using our proposed method.
We present the experimental images of real-world scenario in tasks such as Pick\_place, as shown in Fig. \ref{fig:real-world pick_place} - \ref{fig:real-world shake_mistake}.
Notably, our framework allows flexible task skipping based on real-time conditions.  
In the \texttt{weigh} task, when the beaker is already on the balance, CAPER can bypass this subtask without additional actions.
We additionally analyze common sim-to-real failure cases in Figs.~\ref{fig:real-world weigh_mistake} and \ref{fig:real-world shake_mistake}. 
In the first case, the robot collides with an unseen object because the detector fails to recognize it.
This object has not been included in our YOLOv8 fine-tuning set, leading to a missing detection and an incorrect motion plan. 
In the second case, the robot fails to grasp the target object due to inaccurate depth-based localization from the D415 sensor, resulting in insufficient gripper contact during execution.


\end{document}